\documentclass[journal,twoside,web]{ieeecolor}
\usepackage{tmi}
\usepackage{cite}
\usepackage{amsmath,amssymb,amsfonts}
\usepackage{algorithmic}
\usepackage{graphicx}
\usepackage{booktabs}
\usepackage{textcomp}
\usepackage{subfig}
\usepackage{placeins}
\usepackage{soul}
\usepackage{listings}
\lstdefinestyle{custom}{
    basicstyle=\ttfamily\small, 
    backgroundcolor=\color{white}, 
    frame=single, 
    breaklines=true, 
    rulecolor=\color{black}, 
    commentstyle=\color{green},
    keywordstyle=\color{blue},
    stringstyle=\color{red},
    showstringspaces=false,
}

\def\BibTeX{{\rm B\kern-.05em{\sc i\kern-.025em b}\kern-.08em
    T\kern-.1667em\lower.7ex\hbox{E}\kern-.125emX}}
\begin{document}
\title{From Gaze to Insight: Bridging Human Visual Attention and Vision Language Model Explanation for Weakly-Supervised Medical Image Segmentation}
\author{Jingkun Chen, Haoran Duan, Xiao Zhang,  Boyan Gao, Vicente Grau, Jungong Han\thanks{Corresponding author: Jungong Han}
\thanks{Jingkun Chen, Boyan Gao, and Vicente Grau are with the Department of Engineering Science, University of Oxford, Oxford OX3 7DQ, United Kingdom (e-mail: jingkun.chen; boyan.gao; vicente.grau@eng.ox.ac.uk). Jingkun Chen and Vicente Grau were funded partly by a Novo Nordisk-funded grant.}
\thanks{Jungong Han and Haoran Duan are with the Department of Automation, Tsinghua University, Beijing 100084, China (e-mail: jghan, haoranduan@tsinghua.edu.cn).}
\thanks{Xiao Zhang is with the School of Information Science and Technology, Northwest University, Xi'an  710127, China (e-mail: xiaozhang@nwu.edu.cn).}
}

\maketitle
\begin{abstract}
Medical image segmentation remains challenging due to the high cost of pixel-level annotations for training. In the context of weak supervision, clinician gaze data captures regions of diagnostic interest; however, its sparsity limits its use for segmentation. In contrast, vision-language models (VLMs) provide semantic context through textual descriptions but lack the explanation precision required. Recognizing that neither source alone suffices, we propose a teacher-student framework that integrates both gaze and language supervision, leveraging their complementary strengths. Our key insight is that gaze data indicates ``where" clinicians focus during diagnosis, while VLMs explain ``why" those regions are significant. To implement this, the teacher model first learns from gaze points enhanced by VLM-generated descriptions of lesion morphology, establishing a foundation for guiding the student model. The teacher then directs the student through three strategies: (1) Multi-scale feature alignment to fuse visual cues with textual semantics; (2) Confidence-weighted consistency constraints to focus on reliable predictions; (3) Adaptive masking to limit error propagation in uncertain areas. Experiments on the Kvasir-SEG, NCI-ISBI, and ISIC datasets show that our method achieves Dice scores of 80.78\%, 80.53\%, and 84.22\%, respectively—improving 3–5\% over gaze baselines without increasing the annotation burden. By preserving correlations among predictions, gaze data, and lesion descriptions, our framework also maintains clinical interpretability. This work illustrates how integrating human visual attention with AI-generated semantic context can effectively overcome the limitations of individual weak supervision signals, thereby advancing the development of deployable, annotation-efficient medical AI systems. Code is available at: https://github.com/jingkunchen/FGI.
\end{abstract}


\section{Introduction}

\IEEEPARstart{M}{edical} image segmentation is a critical step in numerous clinical workflows, including disease localization, therapy planning, and longitudinal patient monitoring \cite{azad2024medical}. Despite the advances brought forth by deep learning, especially with fully supervised approaches \cite{chen2021transunet}, the challenge of obtaining large-scale, high-quality pixel-level annotations continues to impede broader adoption. Such annotations often demand extensive domain expertise and can be prohibitively time-consuming in practice \cite{wang2021annotation}.

Efforts to reduce annotation overhead have sparked interest in \emph{weakly supervised} segmentation, with model training steered by coarse labels such as image-level tags, scribbles, bounding boxes, or points \cite{ tajbakhsh2020embracing, cheng2023boxteacher, li2024scribformer, cheng2022pointly, muller2024weakly, zhang2025helpnet}. Although these forms of supervision offer a more efficient alternative to full manual delineation, they frequently lack the precise spatial details necessary to capture subtle pathological boundaries or complex anatomical structures \cite{chen2020adversarial}. Consequently, the community has continued to explore additional weak signals or novel labeling strategies that strike a better balance between annotation economy and segmentation quality.

Recent studies have highlighted the value of \emph{human gaze} data, particularly from clinical experts, as a unique weak supervisory cue \cite{zhong2024weakly, xie2024integrating}. Since medical professionals typically fixate on diagnostically relevant regions, gaze traces naturally indicate pathological areas of interest, potentially serving as a proxy for more detailed annotations. In practice, this modality can be captured using eye-tracking devices during routine clinical assessments, incurring minimal additional cost \cite{duan2025parameter}. However, gaze data also come with inherent challenges: not every fixation corresponds to an abnormality (\emph{e.g.}, due to distractions or exploratory scanning), and the resulting segmentations can be sparse or incomplete, lacking well-defined boundaries \cite{shi2024survey, jiang2024glanceseg}.

Simultaneously, the rapid progress of \emph{vision–language models (VLMs)} has opened new perspectives for medical image analysis \cite{li2023lvit, gao2024boosting, miao2025laser}. By aligning image features with large-scale textual corpora, these models can generate captions or explanations that approximate expert reasoning—such as the expected morphology, size, or location of a lesion \cite{huang2024ligu}. These textual insights can be particularly valuable when spatial cues are incomplete. However, because most VLMs are trained on natural images, they face a significant domain gap when applied to medical imaging. Moreover, semantic ``hallucinations'' are an intrinsic limitation of these models, which typically lack the pixel-level precision required for detailed analysis \cite{peng2023study}. Consequently, directly applying VLM-generated descriptions may not suffice for the robust delineation of subtle abnormalities \cite{chen2024detecting}.

In this paper, to effectively bridge \textbf{``where the expert looked''} with \textbf{``why that region matters''}, we propose a novel \emph{teacher–student} framework that leverages the complementary strengths of human gaze and VLM-based textual explanations. First, we train a \textbf{teacher} model using \emph{high-certainty} gaze pseudo-masks, which are inherently sparse but offer greater reliability. To mitigate coverage gaps, we integrate textual features from a large vision–language model, capturing contextual information such as shape descriptors or boundary details. This synthesis of human fixation and semantic cues guides the teacher to learn refined cross-modal features.

Next, we transfer the refined knowledge to a \textbf{student} model, trained on a more comprehensive yet noisier set of gaze data that may contain spurious fixations. Knowledge distillation is enforced via feature-level alignment—ensuring that the student’s latent representation closely approximates the teacher’s—and a confidence-aware consistency loss that emphasizes regions where the teacher's and student's predictions are both reliable. Moreover, we apply constrained random masking selectively to regions where the teacher and student models exhibit pronounced disagreement. This approach compels the student to harness broader contextual information rather than overfitting to noisy gaze signals, and concurrently attenuates the adverse effects of label noise. We validate our approach on three public benchmarks—Kvasir-SEG (endoscopic polyp segmentation), NCI-ISBI (MRI prostate segmentation), and ISIC (skin lesion segmentation)—demonstrating that the proposed method significantly reduces the performance gap to fully supervised methods while lowering annotation overhead.

\par In summary, our primary contributions are as follows:\begin{itemize}
\item We introduce the first weakly supervised framework for medical image segmentation. It fuses expert gaze patterns with vision–language model outputs to generate clinically precise segmentation cues. These cues are directly optimized via a partial cross‑entropy loss in our teacher network. By doing so, we overcome the limitations of individual weak supervision signals and elevate the VLM from a passive feature extractor to an active supervisory oracle.

\item We propose a novel teacher-student architecture where the teacher model integrates high-confidence gaze annotations with VLM-derived textual embeddings through multi-scale fusion, while the student model learns from noisier gaze data and robust distillation mechanisms.

\item We introduce three key technical innovations: (1) feature-level alignment between visual and textual representations, (2) confidence-aware consistency regularization to focus on reliable predictions, and (3) a disagreement-aware random masking strategy to mitigate label noise.

\item Through extensive experiments on three clinical benchmarks spanning different imaging modalities, we demonstrate significant improvements over existing methods (achieving 3-5\% higher Dice scores) while maintaining the same annotation efficiency, with additional benefits in interpretability through gaze-text-prediction alignment.
\end{itemize}

\section{Related Work}
\label{sec:related_work}

In this section, we review four major research directions pertinent to our work: weakly supervised segmentation methods in medical imaging, the use of human visual attention as a supervisory signal, the application of vision–language models for semantic enrichment, and teacher–student architectures for cross‑modal knowledge transfer. 

\subsection{Weakly Supervised Segmentation in Medical Imaging}
Fully supervised methods, such as UNet \cite{ronneberger2015u} and its variants \cite{chen2021transunet}, have demonstrated impressive results when abundant pixel-level annotations are available \cite{liu2022perturbed, chen2024dynamic}. However, the high cost and time required for dense labeling continue to restrict large-scale data curation in real-world clinical contexts \cite{liu2024translation, liu2024ittakestwo, chen2022semi}. Consequently, weakly supervised approaches have gained traction, exploring alternative supervisory signals such as image-level tags \cite{chen2025unsupervised}, points \cite{cheng2022pointly, wu2023sparsely}, bounding boxes \cite{tian2021boxinst, cheng2023boxteacher}, and scribbles \cite{chen2025addressing, li2024scribformer}.

While these coarse annotations reduce the labeling burden, they often lack the granularity needed to capture subtle anatomical details, leading to performance deficits relative to fully supervised pipelines \cite{kang2025exploring, zhang2023anatomy, chen2020adversarial}. More recent methods investigate point-level annotations, region-level cues, and iterative refinement strategies to mitigate this limitation \cite{wu2024gaze, tajbakhsh2020embracing}. Nonetheless, precise localization of pathologies—particularly those with diffuse or indistinct boundaries—remains challenging. Thus, researchers increasingly seek additional weak signals that can sharpen boundary delineation while retaining annotation efficiency.

\subsection{Human Visual Attention in Medical Imaging}

Eye tracking has emerged as a compelling source of weak supervisory information because it reflects the natural diagnostic focus of clinicians and radiologists \cite{zhong2024weakly, xie2024integrating}. Studies show that fixation patterns often cluster around key pathological regions, effectively serving as coarse, yet task-relevant pointers for potential lesions or abnormalities \cite{ghosh2023automatic}. These gaze maps can guide deep networks toward salient regions, reducing false positives and improving localization \cite{tafasca2024toward}.

Nonetheless, \textbf{noise is inherent in gaze data} \cite{shi2024survey}. Not every fixation is clinically meaningful—some may result from visual search, fatigue, or habit. Additionally, gaze trajectories may only approximate pathology boundaries, leaving the network uncertain about precise contours \cite{jiang2024glanceseg}. To counteract these drawbacks, recent works integrate auxiliary signals or iterative refinement steps, striving to extract more reliable boundary information from limited and occasionally noisy gaze points \cite{song2022learning}. However, purely spatial cues (\emph{i.e.}, ``where experts look'') do not necessarily convey why a region is significant, fueling interest in complementary semantic signals that can bolster or disambiguate gaze-based supervision.

\subsection{Vision-Language Models in Image Analysis}
The advent of large-scale vision–language models has opened new avenues for incorporating high-level semantic context into visual tasks \cite{li2023lvit, gao2024boosting}. Pre-trained transformers \cite{duan2023dynamic} such as CLIP \cite{radford2021learning} have demonstrated robust generalization for natural images, and adaptations for the medical domain (\emph{e.g.}, MedCLIP\cite{wang2022medclip}, BioMedGPT \cite{zhang2024generalist}, MedicalGraphRAG \cite{wu2024medical}) now enable more clinically oriented text generation. These models can produce detailed captions or diagnostic statements that approximate expert reasoning, capturing morphological and contextual insights not easily gleaned from visual features alone \cite{peng2023study,ding2021repvgg, liu2025part}.

Recent endeavors show that textual embeddings can complement spatial cues by highlighting clinically salient concepts—thereby enhancing a network’s interpretive power \cite{huang2024ligu}. Yet a persistent challenge lies in aligning textual semantics with fine-grained spatial details, especially when the text arises from automated generation and may contain inaccuracies (\emph{e.g.}, hallucinated findings \cite{chen2024detecting}). This semantic–spatial misalignment underscores the importance of robust cross-modal fusion mechanisms that can reconcile textual and visual sources of uncertainty.

\subsection{Teacher–Student Architectures and Cross-Modal Knowledge Transfer}
Teacher–student frameworks are purposefully designed to transfer robust, distilled knowledge from a well-trained teacher model to a student model \cite{zhang2024interteach}. In these frameworks, the teacher model provides pseudo-labels guidance through consistency losses \cite{zhang2024cross}. While effective, early approaches often assumed uniformity across all types of data and label sources, thus overlooking variations in annotation quality and domain-specific nuances.

Advanced knowledge distillation methods leverage feature-level alignment and cross‑modal attention \cite{huo2024c2kd} to integrate diverse supervisory signals like points, boxes, and masks by aligning latent representation spaces. However, fixed fusion schedules and uniform weighting can limit the benefits of individual signals—for instance, when a noisy sparse gaze signal or a semantically broad but imprecise automated text signal is used. Therefore, adaptive fusion strategies and selective consistency constraints are needed to ensure robust knowledge transfer in clinical environments.

\section{Methodology}
\label{sec:method}

In this work, we introduce a teacher–student framework that harnesses the complementary strengths of human gaze and vision–language model (VLM) outputs for segmentation in medical imaging. As detailed in Section~\ref{sec:related_work}, gaze data can reliably indicate diagnostically relevant regions but often lack the spatial precision needed for accurate delineation. In contrast, VLMs provide high-level semantic context but may not localize lesions precisely when used directly on medical images. To reconcile these aspects, our method integrates three complementary inputs: (i) \emph{High-confidence Mask}—capturing high-confidence gaze fixations; (ii) \emph{Broad-Coverage Masks}—providing broader yet noisier lesion coverage;  (iii) \emph{VLM-Derived Textual Embeddings}—encoding morphological and semantic attributes.

\begin{figure*}[t]
  \centering
  \includegraphics[width=\textwidth]{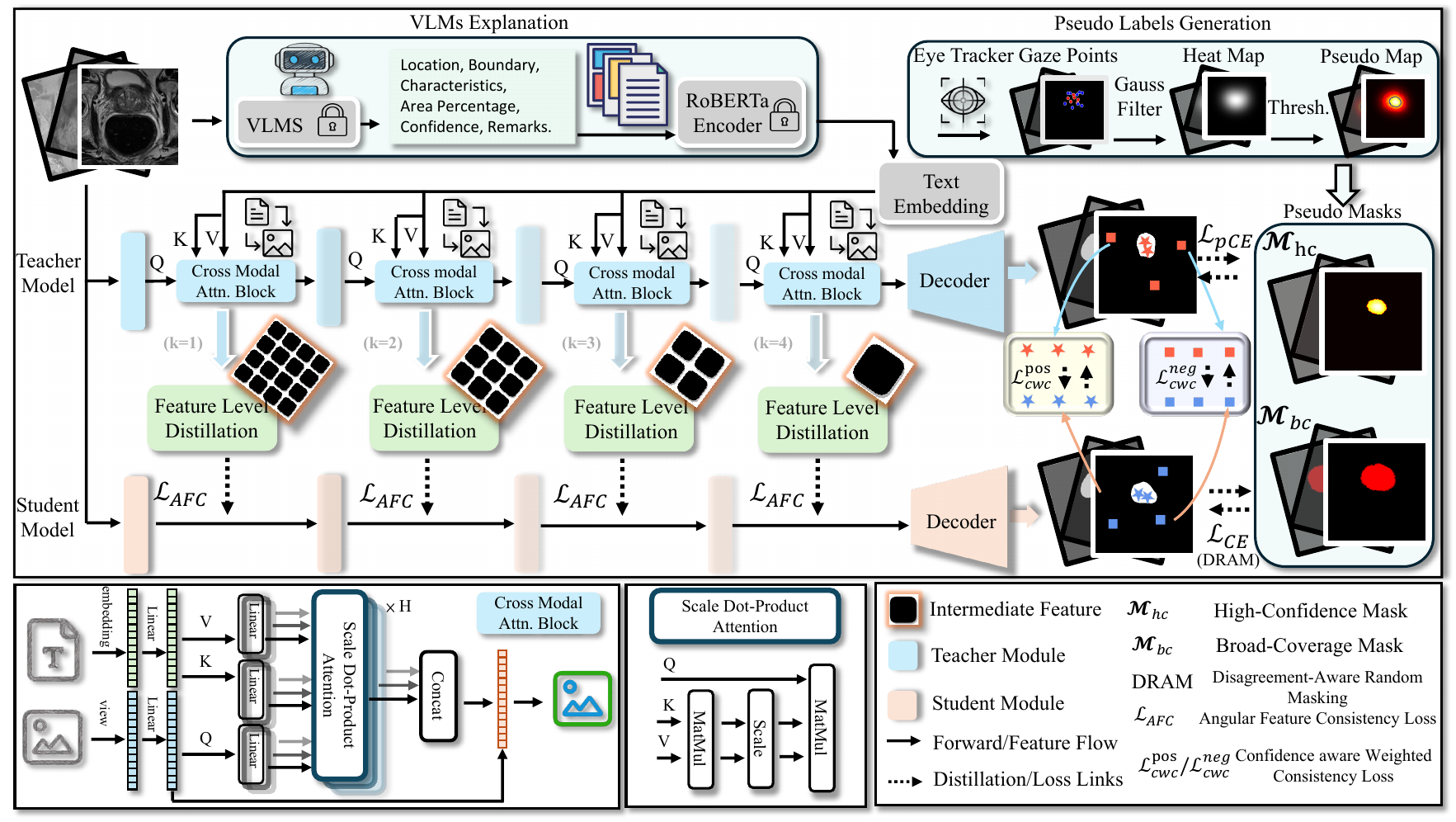}
  \caption{\textbf{Schematic of the proposed gaze- and language-guided teacher–student framework.
A VLM generates a structured explanation (location, boundary, characteristics, area, confidence) whose text embedding is fused with image features in the Teacher via Cross-Modal Attention (Sec.~III-A).
Gaze points are smoothed/thresholded to produce a \emph{high-confidence} mask $M_{hc}$ and a \emph{broad-coverage} mask $M_{bc}$ (Sec.~III-A).
Teacher$\rightarrow$Student \emph{feature-level distillation} operates at stages $k{=}1\ldots4$, with the Angular Feature Consistency loss $\mathcal{L}_{AFC}$ aligning intermediate representations.
Supervision uses $\mathcal{L}_{pCE}(M_{hc})$ and $\mathcal{L}_{CE}(M_{bc})$ (with DARM applied to $M_{bc}$ prior to $\mathcal{L}_{CE}$).
The \emph{Confidence-aware Weighted Consistency loss} $\mathcal{L}_{\mathrm{CWC}}
=\mathcal{L}^{\mathrm{pos}}_{\mathrm{cwc}}+\mathcal{L}^{\mathrm{neg}}_{\mathrm{cwc}}$
is computed on the class-wise logits over $(\Omega_{\mathrm{pos}},\Omega_{\mathrm{neg}})$.}}
  \label{fig:big_pipeline}
\end{figure*}

\begin{figure}[ht]
  \centering
  \includegraphics[width=\linewidth]{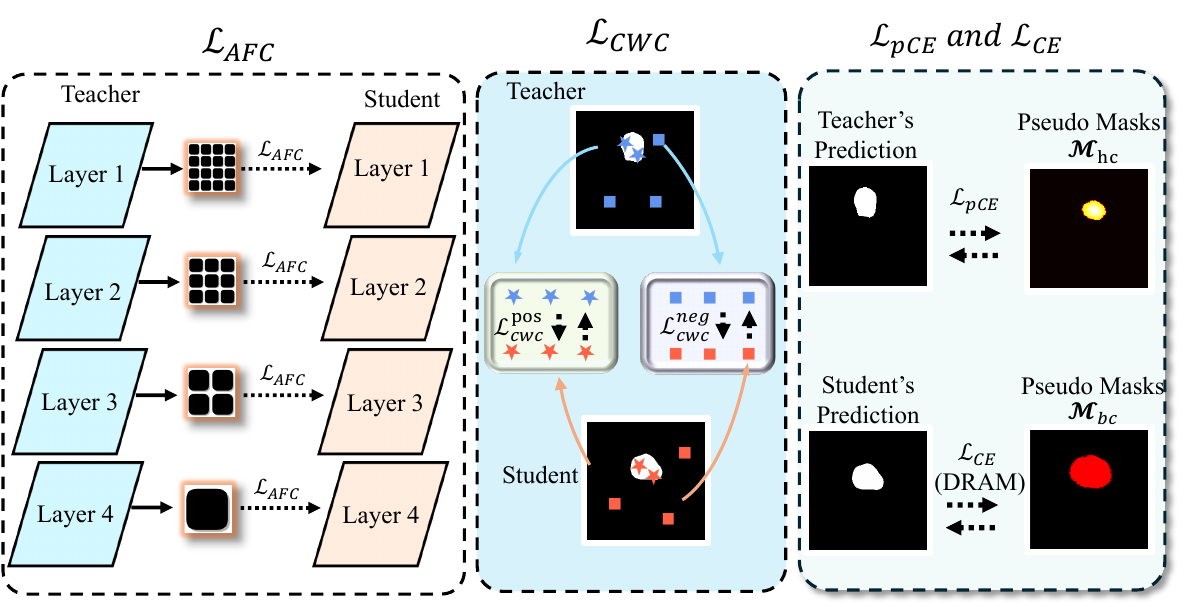} 
  \caption{Overview of the losses used in our framework. Left: Angular Feature Consistency ($\mathcal{L}_{\mathrm{AFC}}$) aligns intermediate features between the teacher and the student. Middle: Confidence-aware Weighted Consistency ($\mathcal{L}_{\mathrm{CWC}}$) regularizes predictions on confident positive/negative regions. Right: Supervision terms---the teacher uses partial cross-entropy ($\mathcal{L}_{\mathrm{pCE}}$) on high-confidence mask ($\mathbf{M}_{\mathrm{hc}}$), while the student uses cross-entropy ($\mathcal{L}_{\mathrm{CE}}$) on broad-coverage masks ($\mathbf{M}_{\mathrm{bc}}$) with disagreement-aware random masking (DRAM).}
  \label{fig:loss}
\end{figure}

As illustrated in Figure~\ref{fig:big_pipeline}, our method begins with the teacher stage, where image features are enhanced with textual cues derived from a vision-language model. A fusion module then uses high-confidence mask to supervise feature refinement via a partial cross-entropy loss, yielding robust cross-modal representations. The teacher model extracts these representations to guide the student network, which is trained on broad-coverage masks via feature-level knowledge distillation combined with a confidence-aware consistency regularization strategy. 
Figure~\mbox{\ref{fig:loss}} summarizes all optimization terms used in our framework: the teacher is supervised by partial cross-entropy ($\mathcal{L}_{\mathrm{pCE}}$) on $\mathbf{M}_{\mathrm{hc}}$; the student is optimized with cross-entropy ($\mathcal{L}_{\mathrm{CE}}$) on $\mathbf{M}_{\mathrm{bc}}$ under disagreement-aware random masking (DRAM); cross-model transfer is enforced via Angular Feature Consistency ($\mathcal{L}_{\mathrm{AFC}}$), and reliability is promoted by Confidence-aware Weighted Consistency ($\mathcal{L}_{\mathrm{CWC}}$).  Next, we provide a detailed description of each module within our framework.

\subsection{Obtaining Gaze Pseudo-Masks and Vision--Language Text}

\paragraph{High-Confidence vs.\ Broad-Coverage Pseudo-Masks}
Building on \cite{zhong2024weakly}, we acquire two sets of gaze-derived masks for each image:
\begin{enumerate}
    \item \textbf{High-Confidence Mask} ($M_{\mathrm{hc}}$): a small, meticulously confirmed masks set capturing core diagnostic fixations, yielding precise but partial annotations.
    \item \textbf{Broad-Coverage Mask} ($M_{\mathrm{bc}}$): a more extensive record covering potential lesion regions, yet containing noise (\emph{e.g.}, ambiguous or off-target gaze).
\end{enumerate}
The high-confidence mask offers near-certain ground truth for a subset of pixels, while the broad-coverage mask—though broader—risks overlabeling. Our method first capitalizes on the high certainty of $M_{\mathrm{hc}}$ in a \emph{teacher} stage, then transitions to $M_{\mathrm{bc}}$ during the \emph{student} stage.

\paragraph{Prompting a Vision--Language Model}
To incorporate domain-level knowledge, we feed each image into a large VLM (\emph{e.g.}, \texttt{doubao-1-5-vision-pro, Chatgpt-4o}) using a structured prompt:
The following structured prompt guides the analysis of abnormal regions (\emph{e.g.}, tumors or polyps) in medical images:

\begin{lstlisting}[style=custom]
  Location: Region position (e.g., upper left)
  Boundary: continuity/clarity/irregularities
  Characteristics: Shape, texture, brightness
  Area Percentage: Estimated area occupied (%)
  Confidence: Score (high/moderate/low)
  Remarks: Additional notes
\end{lstlisting}

The vision–language model returns a structured JSON object detailing key attributes of the suspicious region. Specifically, the JSON includes keys for location, boundary features, morphological characteristics, and estimated area percentage, each governed by predefined constraints. For example, the location key accepts options such as ``upper left'' ``upper right,'' ``lower left,'' or ``lower right.'' The boundary features key is limited to descriptors like ``clear,'' ``irregular,'' or ``ambiguous,'' while the morphological characteristics are restricted to standardized terms (\emph{e.g.}, ``smooth,'' ``spiculated,'' or ``lobulated''). The estimated area percentage is expressed as a numerical value between 0 and 100, representing the proportion of the region occupied by the abnormality. By enforcing these restrictions, the prompt enhances semantic precision and minimizes the risk of generating imprecise or hallucinated data due to domain discrepancies, ensuring that the extracted textual cues are clinically relevant and directly applicable to subsequent stages of data fusion and analysis.

\begin{figure}[!ht]
  \centering
  \includegraphics[width=1\linewidth]{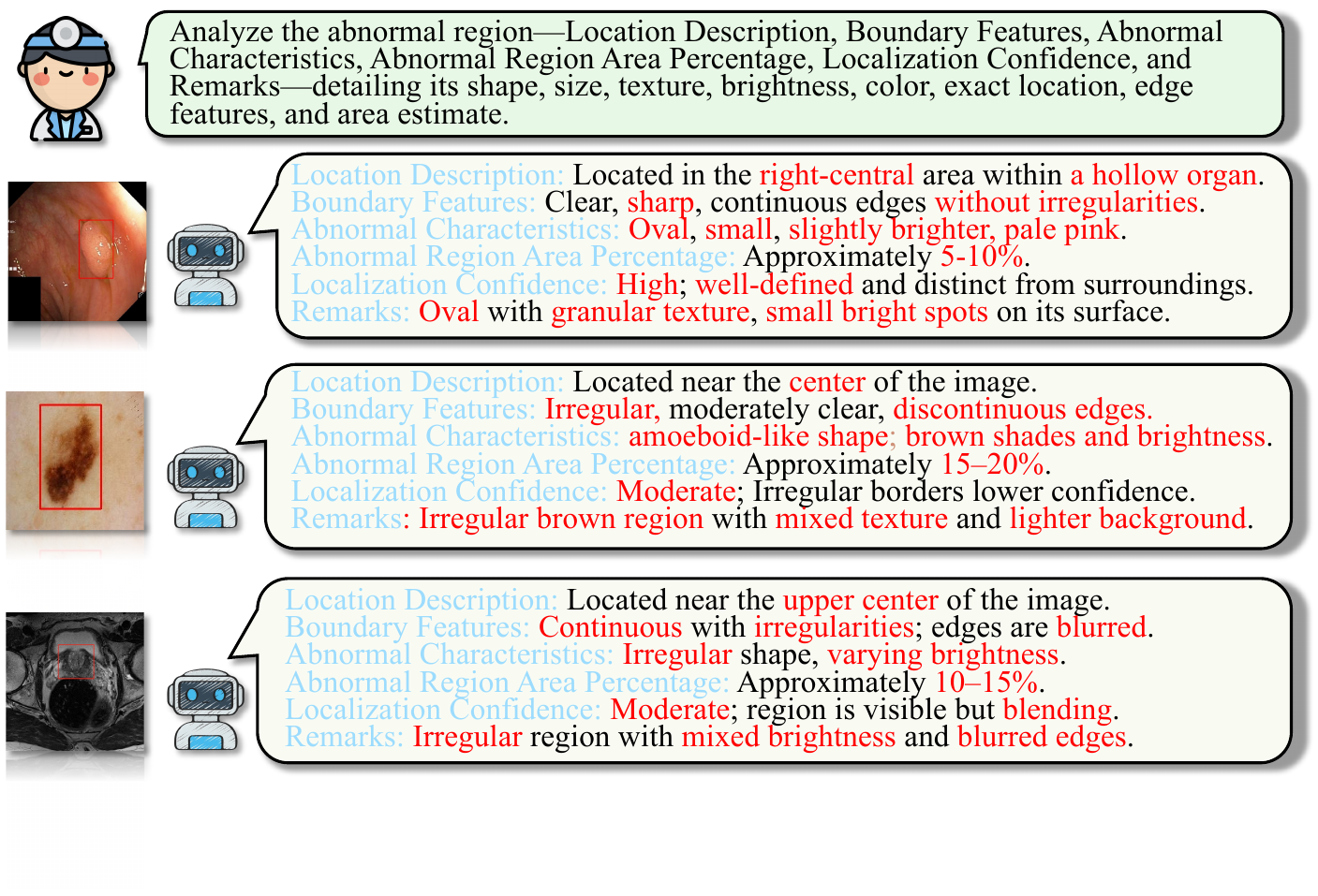} 
  \caption{Schematic illustration of using a large vision–language model to generate fixed-format textual descriptions.}
  \label{fig:vlm_schematic}
\end{figure}

To further illustrate this process, Figure~\ref{fig:vlm_schematic} presents a schematic diagram that outlines how the large language model is employed to generate a fixed-format textual description. In this process, each field of the structured report—covering aspects such as the region's location, boundary quality, morphological characteristics, and estimated area percentage—is first converted into a tokenized representation. This tokenization ensures that every element is normalized and aligned with a predefined set of acceptable descriptors, promoting consistency and reducing variability across outputs.

Subsequently, these tokenized representations are embedded using a large language model (\emph{e.g.}, RoBERTa \cite{liu2019roberta}). The embedding process transforms the structured text into a high-dimensional semantic vector, denoted as $\mathbf{F}_{\mathrm{t}}$. This semantic vector encapsulates comprehensive information regarding the lesion, including detailed morphological descriptions and explicit boundary cues. Additionally, the framework integrates a coarse confidence measure for localizing the abnormal region, aiming to mitigate potential issues arising from the domain gap when applying models primarily trained on natural images to medical imaging tasks.

By constructing $\mathbf{F}_t$ in this manner, the model leverages domain-specific textual insights and facilitates effective fusion with visual features. This enriched semantic representation improves the interpretability and accuracy of downstream data fusion and analysis, ultimately contributing to a robust and reliable segmentation framework.

\paragraph{Multi-Scale Text–Vision Fusion}
We propose a multi-scale fusion module that injects semantic cues from text into visual features using multi-head cross-attention. In practice, this module is applied at each scale (\emph{e.g.}, the different layers of a UNet encoder) to enhance the feature maps with text-guided information.

\textbf{Step 1: Visual Query Preparation.}  
At each scale level \(s\), a visual feature map 
$\mathbf{X}_s \in \mathbb{R}^{H_s \times W_s \times C_v}$
is first flattened into a sequence of tokens and then enriched with positional information. Specifically, the feature map is reshaped to form query vectors:
\begin{equation}
\mathbf{Q}_s = \phi(\mathbf{X}_s) + \mathbf{P}_s,
\end{equation}
where \(\phi(\cdot)\) projects the visual features into a query space, and \(\mathbf{P}_s \in \mathbb{R}^{N \times C_v}\) is a set of learnable positional embeddings (with \(N = H_s\times W_s\)) that inject spatial awareness.

\textbf{Step 2: Textual Key–Value Projection.}  
The text embedding produced by a vision–language model,
\begin{equation}
\mathbf{F}_t \in \mathbb{R}^{L \times C_t},
\end{equation}
is projected into the key and value spaces using two separate linear transformations:
\begin{equation}
\mathbf{K}_s = \psi_k(\mathbf{F}_t),\quad \mathbf{V}_s = \psi_v(\mathbf{F}_t).
\end{equation}
This ensures the text features are compatible with the visual query dimension.

\textbf{Step 3: Multi-Head Cross-Attention.}  
The refined visual queries attend to the text-derived keys and values via a multi-head attention mechanism. For each head, the attention output is computed as:
\begin{equation}
\text{head}_i^k = \sum_j \alpha_{i,j}^k\, \mathbf{v}_{s,j}^k,
\end{equation}
where the attention weights \(\alpha_{i,j}^k\) are given by:
\begin{equation}
\alpha_{i,j}^k = \frac{\exp\Bigl(\frac{\mathbf{q}_i^k \, (\mathbf{k}_{s,j}^k)^\top}{\sqrt{d}}\Bigr)}{\sum_j \exp\Bigl(\frac{\mathbf{q}_i^k \, (\mathbf{k}_{s,j}^k)^\top}{\sqrt{d}}\Bigr)}.
\end{equation}
The outputs of all heads are then concatenated and linearly transformed:
\begin{equation}
\text{MHAtt}(\cdot) = \text{Concat}\bigl(\text{head}^1, \dots, \text{head}^h\bigr)\cdot\mathbf{W}^O,
\end{equation}
with \(\mathbf{W}^O\) being the projection matrix.

\textbf{Step 4: Residual Fusion and Normalization.}  
Finally, the output of the multi-head attention is combined with the original visual queries using a residual connection:
\begin{equation}
\mathbf{X}_s' = \mathbf{X}_s + \lambda\cdot\text{MHAtt}(\mathbf{Q}_s, \mathbf{K}_s, \mathbf{V}_s),
\end{equation}
where \(\lambda\) is a learnable scalar that controls the fusion strength. A layer normalization is applied to ensure stable training.

\paragraph{Partially Cross-Entropy (pCE) Loss}
Since the high-confidence mask \(M_{\mathrm{hc}}\) provides supervision only on a subset of pixels, we employ a \emph{partial cross-entropy} loss defined as:
\begin{equation}
\begin{aligned}
\mathcal{L}_{pCE}(y, \hat{y}) &= -\frac{1}{N}\sum_{i=1}^{N} \sum_{c=1}^{C} y_{i,c}\,\log\!\bigl(\hat{y}_{i,c}\bigr),\\[2pt]
\text{where}\quad y_{i,c} &= 
\begin{cases}
1, & \text{if pixel } i \text{ is labeled as class } c \text{ in } M_{\mathrm{hc}},\\[3pt]
0, & \text{otherwise.}
\end{cases}
\end{aligned}
\end{equation}
Unlabeled pixels are excluded from the loss, allowing the fusion module to leverage text cues for regions with uncertain supervision. These components enable the model to inject semantic information into multi-scale visual features, producing enriched representations that facilitate feature fusion.

\subsection{Student Network: Feature-Level Knowledge Distillation and Confidence-aware Weighted Consistency}

While the teacher focuses on partial but precise information, the student must handle a \emph{broad-coverage} mask $M_{\mathrm{bc}}$, encompassing full lesion coverage (often with noise). We ensure the student maintains the teacher’s cross-modal insights through two key strategies.

\paragraph{Feature-Level Knowledge Distillation}
To transfer the refined representations from the teacher to the student, we enforce feature-level alignment using the \textbf{Angular Feature Consistency Loss (AFC)}, which measures similarity between features via their normalized inner products:
\begin{equation}
\mathcal{L}_{\text{AFC}} = \beta \cdot \frac{1}{L}\sum_{k=1}^L \mathbb{E}\left[1 - \frac{\langle \mathbf{Z}_k, \mathbf{X}'_k \rangle}{\|\mathbf{Z}_k\| \|\mathbf{X}'_k\| + \epsilon}\right], \label{eq:afc}
\end{equation}
where:
\begin{itemize}
\item $\langle \cdot, \cdot \rangle$ denotes the channel-wise inner product,
\item $\|\cdot\|$ represents the $L^2$-norm computed along the channel dimension,
\item $\epsilon=10^{-6}$ ensures numerical stability,
\item $\beta$ is the loss scaling factor (default is 1.0).
\end{itemize}

\paragraph{Confidence-aware Weighted Consistency (CWC) Loss}
To enforce bidirectional consistency in regions where both teacher and student are confident, we propose the Confidence-aware Weighted Consistency (CWC) Loss. In our formulation, the joint confidence of teacher and student predictions is used to define two pixel sets.

\textbf{Confident Positive Regions:}
\begin{equation}
\Omega_{\mathrm{pos}} = \Big\{ p \,\Big|\, 
\begin{aligned}
\max\Bigl(\widehat{\mathbf{Y}}_{\mathrm{teacher}}(p)\Bigr) &\geq \tau_{\mathrm{pos}},\\[1mm]
\max\Bigl(\widehat{\mathbf{Y}}_{\mathrm{student}}(p)\Bigr) &\geq \tau_{\mathrm{pos}}
\end{aligned}
\Big\}.
\label{eq:pos}
\end{equation}

\textbf{Confident Negative Regions:}
\begin{equation}
\Omega_{\mathrm{neg}} = \Big\{ p \,\Big|\, 
\begin{aligned}
\min\Bigl(\widehat{\mathbf{Y}}_{\mathrm{teacher}}(p)\Bigr) &\leq \tau_{\mathrm{neg}},\\[1mm]
\min\Bigl(\widehat{\mathbf{Y}}_{\mathrm{student}}(p)\Bigr) &\leq \tau_{\mathrm{neg}}
\end{aligned}
\Big\}.
\label{eq:neg}
\end{equation}

The CWC loss is then defined as:
\begin{equation}
\mathcal{L}_{\text{CWC}} = 
\underbrace{\frac{1}{|\Omega_{\mathrm{pos}}|}\sum_{p\in\Omega_{\mathrm{pos}}} w(p)\,\ell_{\mathrm{CE}}(p)}_{\text{Positive Consistency}} + 
\underbrace{\frac{1}{|\Omega_{\mathrm{neg}}|}\sum_{p\in\Omega_{\mathrm{neg}}} w'(p)\,\ell_{\mathrm{inv}}(p)}_{\text{Negative Consistency}},
\label{eq:cwc}
\end{equation}
where:
\begin{itemize}
    \item \(w(p) = \max\Bigl(\widehat{\mathbf{Y}}_{\mathrm{teacher}}(p)\Bigr)\) is the positive weight,
    \item \(w'(p) = \min\Bigl(\widehat{\mathbf{Y}}_{\mathrm{teacher}}(p)\Bigr)\) is the negative weight,
    \item \(\ell_{\mathrm{CE}}(p) = -\log \widehat{\mathbf{Y}}_{\mathrm{student}}(y_{\mathrm{teacher}}|p)\) denotes the cross-entropy loss,
    \item \(\ell_{\mathrm{inv}}(p) = -\log\Bigl(1 - \widehat{\mathbf{Y}}_{\mathrm{student}}(1-y_{\mathrm{teacher}}|p)\Bigr)\) is the inverse loss.
\end{itemize}
We set the thresholds as \(\tau_{\mathrm{pos}}=0.8\) and \(\tau_{\mathrm{neg}}=0.2\). This loss ensures that, in regions where both teacher and student exhibit high confidence, their predictions are mutually aligned, thereby enhancing overall consistency and robustness.

\paragraph{Comprehensive Cross-Entropy \& Random Masking}
Finally, we include a standard cross-entropy term over $M_{\mathrm{bc}}$. Splitting it across two lines to avoid overflow:
\begin{align}
\label{eq:ce_comp}
\mathcal{L}_{\mathrm{CE}}&(M_{\mathrm{bc}}) = -\sum_{p}
\Bigl[
M_{\mathrm{bc}}(p)\,\log\!\bigl(\widehat{Y}_{\mathrm{student}}(p)\bigr)
\\
&~+~
\bigl(1 - M_{\mathrm{bc}}(p)\bigr)\,\log\!\bigl(1-\widehat{Y}_{\mathrm{student}}(p)\bigr)
\Bigr].
\notag
\end{align}

Since some pixels in the broad-coverage mask $M_{bc}$ can be unreliable, we apply Disagreement-Aware Random Masking (DARM) before computing $\mathcal{L}_{CE}(M_{bc})$: let $\widehat{\mathbf{Y}}_{\mathrm{T}}, \widehat{\mathbf{Y}}_{\mathrm{S}} \in [0,1]^C$ denote the Teacher/Student supervision, and define the high-disagreement set $\mathcal{D}=\{\,p \mid \lVert \widehat{\mathbf{Y}}_{\mathrm{T}}(p)-\widehat{\mathbf{Y}}_{\mathrm{S}}(p)\rVert_{\infty} \ge \tau_{\mathrm{dis}} \,\}$ with $\tau_{\mathrm{dis}}\in(0,1)$. At each iteration, we sample non-overlapping $s\times s$ patches $\mathcal{P}\subseteq\mathcal{D}$ at rate $\rho$ and form a masked supervision $\tilde{M}_{bc} = M_{bc}\odot (1-\mathbf{1}_{\mathcal{P}})$, where $\odot$ denotes element-wise multiplication and $\mathbf{1}_{\mathcal{P}}(p)=1$ if $p$ lies in any selected patch (else $0$). The cross-entropy is then computed on $\tilde{M}_{bc}$, i.e., $\mathcal{L}_{CE}(\tilde{M}_{bc})$, which selectively occludes inconsistent supervision, reduces the impact of noisy annotations, and encourages reliance on more reliable contextual cues. Combining the terms, the Student objective is.
\begin{equation}
\label{eq:student_loss}
\mathcal{L}_{\mathrm{student}}
=
\mathcal{L}_{CE}(M_{bc})
+
\lambda_{\mathrm{AFC}}\,\mathcal{L}_{AFC}
+
\lambda_{\mathrm{CWC}}\,\mathcal{L}_{\mathrm{CWC}}.
\end{equation}
Here, $\mathcal{L}_{CE}(M_{bc})$ supervises the mask branch using the broad-coverage pseudo-label
(\emph{implemented on its DARM-masked variant} $\tilde M_{bc}$);
$\mathcal{L}_{AFC}$ aligns Teacher–Student \emph{intermediate features} at fusion stages $k{=}1\ldots4$;
and $\mathcal{L}_{\mathrm{CWC}}$ imposes \emph{confidence-aware consistency} on the \emph{class-wise logits} over $(\Omega_{\mathrm{pos}},\Omega_{\mathrm{neg}})$.
The weights $\lambda_{\mathrm{AFC}}$ and $\lambda_{\mathrm{CWC}}$ balance the respective contributions.
For completeness, the Teacher is optimized with the Partially Cross-Entropy loss $\mathcal{L}_{pCE}(M_{hc})$.

\subsection{Rationale and Benefits}

Our approach capitalizes on:
\begin{itemize}
    \item \textbf{High-Confidence Mask}: Reliable yet partial gaze annotations that can be used with partial cross-entropy, preventing over-penalization in unlabeled areas.
    \item \textbf{Textual Explanation}: Structured reports describing ``Location'', ``Boundary'', ``Area'', and ``Localization Confidence'', providing morphological cues and an assessment of how confidently the VLM localizes the lesion.
    \item \textbf{Broad-Coverage Masks}: Wider lesion coverage, albeit noisier, refined through teacher distillation and high-confidence consistency checks.
\end{itemize}
By first molding a teacher with robust cross-modal reasoning, then scaling it up via a student that confronts broader but riskier labels, we marry the \emph{precision} of minimal gaze with the \emph{contextual awareness} of domain text and the \emph{coverage} of comprehensive fixation data. This significantly reduces the need for full manual annotation while preserving interpretability and accuracy—effectively bridging \textbf{``where the expert looked''} with \textbf{``why that region matters.''}

\section{Experiments}
\subsection{Datasets and Evaluation Protocol}

Following the evaluation protocol in~\cite{zhong2024weakly}, we conduct experiments on three medical image segmentation benchmarks. For Kvasir-SEG~\cite{jha2020kvasir} (900 training/100 test endoscopic images) and NCI-ISBI~\cite{nci-isbi} (789 training/117 test MRI scans), we adopt the standard Dice coefficient and annotation time (AT) as evaluation metrics. 

For ISIC~\cite{codella2019skin} (800 training/200 test dermoscopic images), we simulate gaze annotations following the generation rules in~\cite{zhong2024weakly} and employ four complementary metrics: Dice Similarity Coefficient (Dice), mean Intersection-over-Union (mIoU), 95\% Hausdorff Distance (HD95), and Average Surface Distance (ASD). All metrics are computed using standardized evaluation protocols to ensure comparability.

\subsection{Implementation Details}

Our framework comprises three key components. The vision-language module combines Doubao-1.5-Vision-Pro for generating anatomical descriptors with a RoBERTa-based text encoder that produces 768-dimensional embeddings. The teacher model employs a UNet architecture to process gaze annotations through cross-modal feature fusion. The student model uses a lightweight UNet, trained with Angular Feature Consistency Loss (\(\lambda_{\text{AFC}}=0.1\)) for attention alignment and a Confidence-aware Consistency Weight loss (\(\lambda_{\text{CWC}}=1.0\)) for prediction consistency, where the consistency loss is ramped up (i.e., warmed up) during training.

All experiments are conducted on NVIDIA RTX6000ada 48GB GPU using Adam optimizer with initial learning rate $1\times10^{-2}$ and cosine decay schedule. Hyperparameters follow~\cite{zhong2024weakly}. Each configuration is evaluated with three random seeds to ensure statistical reliability.

\subsection{Performance on Kvasir-SEG (Endoscopic Polyp Segmentation)}

As shown in Table~\ref{tab:kvasirseg}, our framework achieves a Dice score of 80.78\% on the Kvasir-SEG dataset, significantly outperforming standard gaze-based weak supervision (70–78\%) and even surpassing methods that rely on bounding-box or scribble annotations (65–77\%). These improvements are particularly evident for smaller or irregularly shaped polyps, where the integration of textual descriptors (\emph{e.g.}, ``lateral edge roughness," ``location") is crucial, as visualized in Figure~\ref{fig:vis_methods}.

Endoscopic polyp segmentation is inherently challenging due to ambiguous lesion boundaries arising from complex textures, partial occlusions, and subtle protrusions. To address these challenges, our approach leverages structured text embeddings from the vision-language model—such as ``semi-regular oval shape" or ``mildly protruding boundary"—to resolve inconsistencies in the gaze-derived weak labels. This effective fusion of visual and textual cues not only boosts overall segmentation performance but also enhances robustness in diagnostically critical edge cases.

\begin{table}[!ht]
  \centering
  \caption{Quantitative comparison on the Kvasir-SEG dataset. Underline indicates the best result among weakly supervised methods. AT denotes annotation time.}
  \renewcommand{\arraystretch}{1.15}
  \setlength{\tabcolsep}{4mm}
  \begin{tabular}{l c c c}
    \toprule
    \textbf{Method} & \textbf{Sup.} & \textbf{mDice (\%)} & \textbf{AT}\\
    \midrule
    UNet           & Full & $82.12 \pm 1.11$ & 18.7 hrs\\
    nnUNet        & Full & $85.37 \pm 0.48$ & 18.7 hrs\\
    \midrule
    BoxInst     & Box & $65.72 \pm 2.97$ & 3.1 hrs\\
    BoxTeacher  & Box & $73.33 \pm 1.30$ & 3.1 hrs\\
    \midrule
    PointSup   & Point & $73.05 \pm 1.64$ & 4.8 hrs\\
    AGMM          & Point & $75.57 \pm 0.84$ & 4.8 hrs\\
    \midrule
    AGMM         & Scribble & $67.23 \pm 1.02$ &2.6 hrs\\
    CycleMix   & Scribble & $76.43 \pm 0.65$ &2.6 hrs\\
    ShapePU     & Scribble & $77.26 \pm 0.73$ &2.6 hrs\\
    ScribFormer  & Scribble & $75.69 \pm 0.48$ & 2.6 hrs\\
    \midrule
    UNet          & Gaze & $73.74 \pm 0.94$& 2.2 hrs \\
    TransUNet  & Gaze & $70.38 \pm 0.86$ & 2.2 hrs\\
    nnUNet        & Gaze & $74.42 \pm 0.92$ & 2.2 hrs\\
    GazeMedSeg  & Gaze & $77.80 \pm 1.02$ & 2.2 hrs\\
    \underline{Ours}             & Gaze & \underline{$80.78 \pm 0.11$} & \underline{2.2 hrs}\\
    \bottomrule
  \end{tabular}
  \label{tab:kvasirseg}
\end{table}

\subsection{Performance on NCI-ISBI (MRI Prostate Segmentation)}
As shown in Table~\ref{tab:nciisbi}, our framework achieves a Dice score of 80.53\% on the NCI-ISBI dataset, outperforming gaze-based alternatives by 3-5 percentage points. This represents a significant improvement over direct use of sparse gaze annotations (74-77\% Dice) with conventional segmentation architectures (\emph{e.g.}, TransUNet or nnUNet). The gains demonstrate that: (1) text embeddings (\emph{e.g.}, ``irregular boundary," ``mucosal protrusion") effectively complement incomplete gaze maps in the teacher model, and (2) our confidence-aware consistency mechanism successfully filters noise from fixation data in the student model.

Notably, while fully supervised approaches (\emph{e.g.}, nnUNet) achieve marginally higher performance (81-82\% Dice), they require approximately 8× more annotation time (~18 hours vs. our 2.2 hours of gaze collection). This favorable trade-off - near state-of-the-art accuracy with dramatically reduced labeling burden - highlights the practical advantage of combining gaze signals with vision-language reasoning for medical image segmentation.

\begin{table}[!ht]
  \centering
  \caption{Quantitative comparison on the NCI-ISBI dataset. Underline indicates the best result among weakly supervised methods. AT denotes annotation time.}
  \renewcommand{\arraystretch}{1.15}
  \setlength{\tabcolsep}{4mm}
  \begin{tabular}{l c c c}
    \toprule
    \textbf{Method} & \textbf{Sup.} & \textbf{mDice (\%)} & \textbf{AT} \\
    \midrule
    UNet \cite{ronneberger2015u}               & Full & $80.58 \pm 0.48$ & 18.7 hrs \\
    nnUNet \cite{isensee2021nnu}           & Full & $81.54 \pm 0.45$ & 18.7 hrs \\
    \midrule 
    BoxInst \cite{tian2021boxinst}         & Box & $73.78 \pm 1.15$ & 3.1 hrs \\
    BoxTeacher \cite{cheng2023boxteacher}   & Box & $75.60 \pm 1.15$ & 3.1 hrs \\
    \midrule
    PointSup \cite{cheng2022pointly}       & Point & $73.46 \pm 4.71$ & 4.8 hrs \\
    AGMM \cite{wu2023sparsely}               & Point & $73.86 \pm 1.26$ & 4.8 hrs \\
    \midrule
    AGMM \cite{wu2023sparsely}               & Scribble & $72.70 \pm 1.03$ & 2.6 hrs \\
    CycleMix \cite{zhang2022cyclemix}       & Scribble & $73.41 \pm 1.09$ & 2.6 hrs \\
    ShapePU \cite{zhang2022shapepu}         & Scribble & $73.06 \pm 1.18$ & 2.6 hrs \\
    ScribFormer \cite{li2024scribformer} & Scribble & $74.31 \pm 1.29$ & 2.6 hrs \\
    \midrule
    UNet \cite{ronneberger2015u}               & Gaze & $74.75 \pm 1.58$ & 2.2 hrs \\
    TransUNet \cite{chen2021transunet}     & Gaze & $75.46 \pm 1.20$ & 2.2 hrs \\
    nnUNet \cite{isensee2021nnu}           & Gaze & $77.20 \pm 1.03$ & 2.2 hrs \\
    GazeMedSeg \cite{zhong2024weakly} & Gaze & $77.64 \pm 0.57$ & 2.2 hrs \\
    \underline{Ours}                 & Gaze & \underline{$80.53 \pm 0.49$} & \underline{2.2 hrs} \\
    \bottomrule
  \end{tabular}
  \label{tab:nciisbi}
\end{table}

\subsection{Performance on ISIC (Skin Lesion Segmentation)}
Table~\ref{tab:isic2018} summarizes the segmentation performance on the ISIC dataset, a challenging dermoscopic dataset characterized by high variability in lesion appearance (\emph{e.g.}, color, texture) and imaging conditions (\emph{e.g.}, lighting, hair occlusions). Fully supervised methods, such as UNet and nnUNet, achieve high performance with mDice scores of 88.17\% and 88.96\%, respectively, reflecting the benefits of dense pixel-level annotations. In contrast, our proposed method, which leverages a teacher–student framework to integrate sparse gaze annotations and vision–language derived textual cues, obtains a mean Dice score of 84.22\%.

Notably, compared with other gaze-based approaches (\emph{e.g.}, UNet, TransUNet, nnUNet, and GazeMedSeg), our method consistently outperforms them across multiple metrics. The higher mIoU and improved boundary metrics—HD95 and ASD—demonstrate that our framework not only enhances overall segmentation accuracy but also achieves better boundary delineation and consistency. 

\begin{table}[!ht]
  \centering
  \caption{Quantitative comparison on the ISIC dataset. Underline indicates the best result among gaze methods.}
  \small
  \setlength{\tabcolsep}{2mm}
  \renewcommand{\arraystretch}{0.95}
  \resizebox{\columnwidth}{!}{%
    \begin{tabular}{l c c c c c}
      \toprule
      \textbf{Method} & \textbf{Sup.} & \textbf{mIoU (\%)} & \textbf{mDice (\%)} & \textbf{HD95} & \textbf{ASD} \\
      \midrule
      UNet   & Full &  $80.97\pm 0.05$ & $88.17\pm0.10$ & $7.96\pm 0.34$ & $3.22\pm 0.10$  \\
      nnUNet & Full &  $81.98\pm 0.29$ & $88.96\pm0.18$ & $7.70\pm 0.20$ & $2.98\pm 0.09$ \\
      \midrule
      UNet       & Gaze &  $73.28\pm 0.61$ & $82.95\pm045$ & $11.65\pm 0.37$ & $5.35\pm 0.19$ \\
      TransUNet  & Gaze &  $69.61\pm 0.78$ & $80.05\pm0.67$ & $13.90\pm 0.27$ & $ 6.23\pm 0.22$ \\
      nnUNet     & Gaze &  $72.38\pm 0.70$ & $82.27\pm0.50$ & $11.87\pm 0.14$ & $5.60\pm 0.08$ \\
      GazeMedSeg & Gaze &  $73.38\pm 0.07$ & $82.91\pm0.02$ & \underline{$10.35\pm 0.47$} & $4.64\pm 0.04$ \\
      \underline{Ours}  & Gaze &  \underline{$75.10\pm 0.21$} & \underline{$84.22\pm0.17$} & $10.76\pm 0.44$ & \underline{$4.21\pm 0.09$} \\
      \bottomrule
    \end{tabular}
  }
  \label{tab:isic2018}
\end{table}

Our experiments show that combining gaze inputs with semantic descriptors significantly boosts segmentation performance by capturing both spatial fixations and contextual morphological cues. The teacher–student framework leverages a teacher model trained on fewer, gaze annotations to pinpoint salient regions and refine boundary details through textual insights, which are then distilled into the student. Additionally, consistency and random masking mitigate noise in gaze pseudo-masks and misinterpretations in language outputs by enforcing alignment in reliably predicted areas and preventing overfitting to uncertain fixations. This strategy achieves substantial gains with a markedly reduced annotation burden compared to fully supervised pipelines, making it an appealing and scalable solution for clinical practice.

\subsection{Ablation Study}

We conduct extensive ablation studies on the Kvasir-SEG dataset to quantify the contribution of each module and loss component in our framework. In the following, we present multiple tables reporting the Dice coefficient (in \%) under different ablation settings.

\subsubsection{Effect of Vision-Language Evaluation Prompt}
To assess the impact of the auxiliary evaluation prompt on the quality of the vision-language model’s output, we compared three settings: Random Prompt, Blank Prompt, and the full VLM Prompt. As shown in Table~\ref{tab:abl_prompt}, the VLM Prompt achieves a Dice score of $80.78\pm0.11\%$, which is notably higher than the Random Prompt’s $78.82\pm1.29\%$ and the Blank Prompt’s $79.76\pm0.16\%$. This result demonstrates that ensuring the accuracy of the textual description is critical for obtaining robust semantic embeddings.

\begin{table}[!ht]
  \centering
  \caption{Effect of Vision-Language Evaluation Prompt on Kvasir-SEG. The table compares three prompt settings: Random Prompt, Blank Prompt, and VLM Prompt.}
  \renewcommand{\arraystretch}{1.15}
  \setlength{\tabcolsep}{8mm}
  \begin{tabular}{l c}
    \toprule
    \textbf{Variant} & \textbf{Dice (\%)} \\
    \midrule
    
    Random Prompt         & $78.82 \pm 1.29$ \\

    Blank Prompt         & $79.76 \pm 0.16$ \\

    VLM Prompt            & \underline{$80.78 \pm 0.11$} \\
    \bottomrule
  \end{tabular}
  \label{tab:abl_prompt}
\end{table}

\subsubsection{Effect of Multi-Scale Text-Vision Fusion Strategy}  
To evaluate the impact of our proposed multi-scale text-vision fusion strategy in the teacher model, we compare two fusion methods. In our approach, image and text features are refined through multi-head cross-attention, and the resulting features are integrated via element-wise summation. For comparison, we implement a simpler variant that fuses the refined features by concatenation. As shown in Table~\ref{tab:abl_attention_fusion}, the multi-scale text-vision fusion strategy achieves a Dice score of $80.78\pm0.11\%$, significantly outperforming the concatenation variant (Dice score of $80.02\pm0.36\%$). These results indicate that our fusion mechanism more effectively aligns the modalities and improves segmentation performance.

\begin{table}[!ht]
  \centering
  \caption{Effect of attention fusion variants on Kvasir-SEG. “Multi-Scale Text-Vision Fusion” fuses visual and textual features using multi-head cross-attention followed by element-wise summation, while “Concat Fusion” fuses the refined features by concatenation.}
  \renewcommand{\arraystretch}{1.15}
  \setlength{\tabcolsep}{8mm}
  \begin{tabular}{l c}
    \toprule
    \textbf{Variant} & \textbf{Dice (\%)} \\
    \midrule
    Concat Fusion Variant & $80.02 \pm 0.36$ \\
    Multi-Scale Text-Vision Fusion & \underline{$80.78 \pm 0.11$} \\
    \bottomrule
  \end{tabular}
  \label{tab:abl_attention_fusion}
\end{table}

\subsubsection{Effect of Loss Components in the Student Model}
To investigate the contribution of loss components in the student model, we conduct experiments by removing the Angular Feature Consistency loss ($\mathcal{L}_{AFC}$) and the confidence-aware Weighted Consistency loss ($\mathcal{L}_{CWC}$) as detailed in Section~\ref{sec:method}.
\begin{enumerate} 

\item \textbf{Full:} Both $\mathcal{L}_{AFC}$ and $\mathcal{L}_{CWC}$ are used. 

\item \textbf{w/o Angular Feature Consistency:} Set $\lambda_{AFC}=0$. 

\item \textbf{w/o Confidence-Aware Consistency:} Set $\lambda_{CWC}=0$.

\item \textbf{w/o Both:} Both loss components are removed. \end{enumerate}

Table~\ref{tab:abl_student_loss_compact} reports the corresponding Dice scores. As shown, the removal of either loss component leads to a performance drop, with the omission of both resulting in the most significant decrease. These results confirm that both $\mathcal{L}_{AFC}$ and $\mathcal{L}_{CWC}$ are essential for effective cross-modal knowledge transfer from the teacher to the student model.

\begin{table}[!ht]
  \centering
  \caption{Ablation study on student loss components (Kvasir-SEG). \checkmark: included; $\times$: removed. Dice scores in \%.}
  \renewcommand{\arraystretch}{1.0}
  \setlength{\tabcolsep}{3mm}
  \begin{tabular}{lccc}
    \toprule
    \textbf{Variant} & \textbf{Angular} & \textbf{Consistency} & \textbf{Dice (\%)} \\
    \midrule
     
     w/o Both         & $\times$   & $\times$   & $ \text{75.45} \pm \text{0.24}$ \\

    w/o CWC & \checkmark & $\times$   & $ \text{75.99} \pm \text{0.26}$ \\

    w/o AFC      & $\times$   & \checkmark & $\text{79.61}\pm\text{0.28}$\\

    w Both       & \checkmark & \checkmark & \underline{$ \text{80.78} \pm \text{0.11}$} \\
 
    \bottomrule
  \end{tabular}
  \label{tab:abl_student_loss_compact}
\end{table}

\begin{figure*}[!t]
  \centering
  \includegraphics[width=1\textwidth]{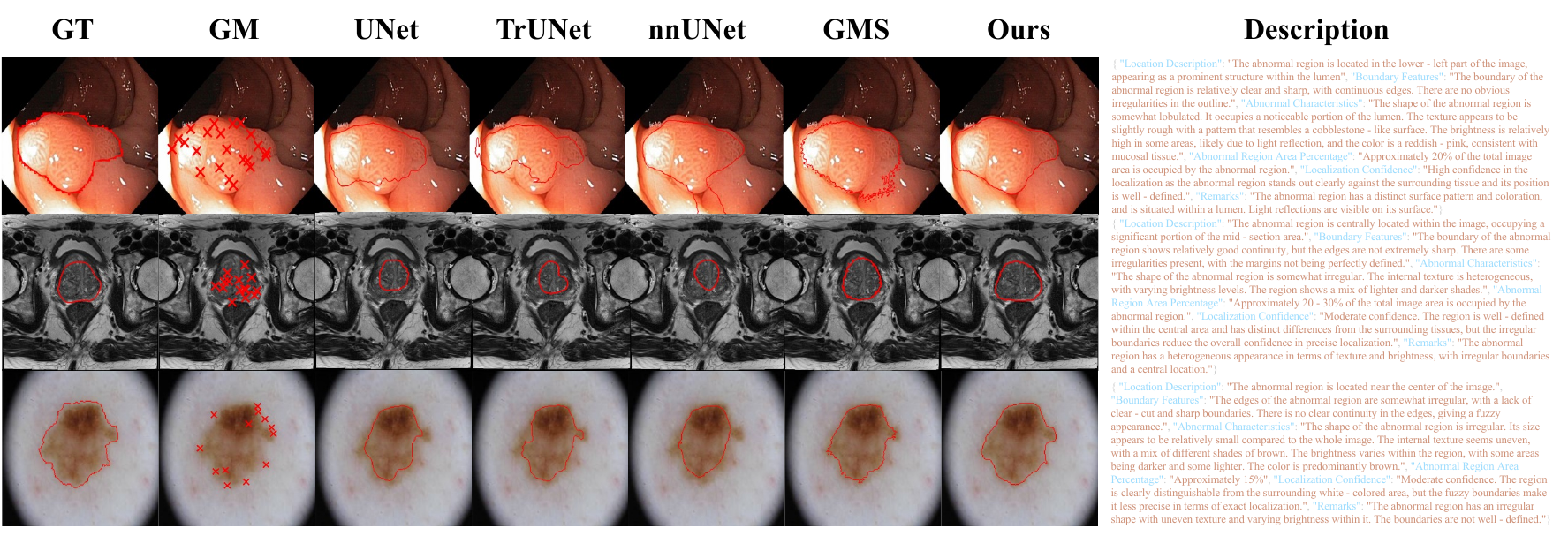}
  \caption{\textbf{Qualitative comparison on three benchmarks.
Rows correspond to Kvasir-SEG, NCI-ISBI, and ISIC. 
Columns (left to right): Ground Truth (GT), Gaze Map (GM), UNet, TransUNet (TrUNet), nnUNet, GazeMedSeg (GMS), Ours, and the VLM-derived \emph{Description}. 
Red overlays indicate gaze fixations and reference boundaries; predicted masks are rendered as semi-transparent overlays for visual comparison. 
Across representative cases, our method yields more complete lesion coverage and sharper boundary adherence while reducing spurious activations in distractor regions.}}
  \label{fig:vis_methods}
\end{figure*}

\subsubsection{Effect of Disagreement-Aware Random Masking}
To mitigate the negative effects of noisy gaze annotations, our method employs a Disagreement-Aware Random Masking (DARM) strategy. In regions where the teacher and student masks exhibit strong disagreement, we randomly zero out local patches—encouraging the student model to rely on broader contextual cues rather than overfitting to unreliable fixation signals, while simultaneously mitigating the adverse effects of label noise. To assess the impact of DARM, we compare the full model (with DARM) against a variant without masking. As shown in Table~\ref{tab:abl_random_masking}, disabling DARM results in a significant drop in the Dice coefficient, confirming that DARM effectively alleviates the influence of inconsistent annotations and improves overall segmentation performance.

\begin{table}[!ht]
  \centering
  \caption{Impact of the disagreement-aware random masking on segmentation performance. Dice scores are reported in \%.}
  \renewcommand{\arraystretch}{1.15}
  \setlength{\tabcolsep}{8mm}
  \begin{tabular}{l c}
    \toprule
    \textbf{Variant} & \textbf{Dice (\%)} \\
    \midrule
    
    w/o DARM             & $ \text{76.21} \pm \text{0.40}$ \\

    w DARM  & \underline{$ \text{80.78} \pm \text{0.11}$} \\
    \bottomrule
  \end{tabular}
  \label{tab:abl_random_masking}
\end{table}

\subsubsection{Effect of Cross-Modal Aggregation Depth}
In our teacher model, the cross-modal aggregation module integrates spatial features with textual embeddings across fusion stages. In this module, Stage 1 aggregates the lowest-level visual features, and each subsequent stage fuses these features with those from a higher level. To investigate the impact of fusion depth, we vary the number of stages from 1 to 4 while keeping all other settings constant. As shown in Table~\ref{tab:abl_pcar_depth}, the Dice score improves with an increasing number of fusion stages, with the 4-stage configuration achieving the best performance.
\begin{table}[!ht]
  \centering
  \caption{Effect of the number of fusion stages in the cross-modal aggregation module on segmentation performance. Dice scores are reported in \%.}

  \renewcommand{\arraystretch}{1.15}
  \setlength{\tabcolsep}{8mm}
  \begin{tabular}{l c}
    \toprule
    \textbf{Fusion Stages} & \textbf{Dice (\%)} \\
    \midrule
    1 Stage  & $ \text{80.05} \pm \text{0.49}$ \\

    2 Stages & $ \text{80.10} \pm \text{0.11}$ \\

    3 Stages & $ \text{80.59} \pm \text{0.44}$ \\

    4 Stages & $ \underline{\text{80.78} \pm \text{0.11}}$ \\

    \bottomrule
  \end{tabular}
  \label{tab:abl_pcar_depth}
\end{table}

\subsubsection{Effect of Confidence Thresholds on Selecting Confident Regions}

To investigate the impact of the confidence thresholds used for selecting reliable regions in the Confidence-Aware Consistency Loss, we conduct an ablation study on the Kvasir-SEG dataset. We compare four threshold pairs: \(\tau_{\mathrm{neg}}=0.1,\, \tau_{\mathrm{pos}}=0.9\); \(\tau_{\mathrm{neg}}=0.2,\, \tau_{\mathrm{pos}}=0.8\); \(\tau_{\mathrm{neg}}=0.3,\, \tau_{\mathrm{pos}}=0.7\); and \(\tau_{\mathrm{neg}}=0.4,\, \tau_{\mathrm{pos}}=0.6\).

\begin{table}[!ht]
  \centering
  \renewcommand{\arraystretch}{1.15}
  \setlength{\tabcolsep}{8mm}
  \caption{Effect of different threshold pairs ($\tau_{\mathrm{neg}}$, $\tau_{\mathrm{pos}}$) on segmentation performance. Dice scores are reported in \%.}
   \begin{tabular}{l c}
    \toprule
    \textbf{Hyperparameters} & \textbf{Dice (\%)} \\
    \midrule
    $\tau_{\mathrm{neg}} = 0.1,\quad \tau_{\mathrm{pos}} = 0.9$ & $80.54 \pm 0.33$ \\
    $\tau_{\mathrm{neg}} = 0.2,\quad \tau_{\mathrm{pos}} = 0.8$ & \underline{$80.78 \pm 0.11$} \\
    $\tau_{\mathrm{neg}} = 0.3,\quad \tau_{\mathrm{pos}} = 0.7$ & $80.58 \pm 0.14$ \\
    $\tau_{\mathrm{neg}} = 0.4,\quad \tau_{\mathrm{pos}} = 0.6$ & $80.11 \pm 0.14$ \\

    \bottomrule
  \end{tabular}
  \label{tab:abl_thresholds}
\end{table}
Table~\ref{tab:abl_thresholds} summarizes the Dice scores obtained with these configurations. As shown, the default threshold configuration yields the highest performance, suggesting that setting pixels with a maximum prediction confidence above 0.8 as positive and those below 0.2 as negative effectively balances the supervision by filtering out low-confidence predictions while retaining sufficient supervision signals.

\section{Visualization}

Figure~\ref{fig:vis_methods} presents a comprehensive qualitative comparison of segmentation results using gaze-based annotations. The multi-row, multi-column figure illustrates results from three datasets: KSEG (Kvasir-SEG), NCI (NCI-ISBI) and ISIC. Each row corresponds to a dataset, and the columns from left to right show the Ground Truth mask, the Gaze Map (GM) (note that for the NCI dataset only pseudo labels were provided so a schematic is shown), segmentation outputs from UNet, TransUNet (TrUNet), and nnUNet, the GazeMedSeg (GMS) results and our Proposed Method (Ours), as well as a brief Description.

These visualizations clearly demonstrate that our proposed method effectively integrates both visual and textual information. The approach combines gaze-based annotations with derived pseudo-labels to produce segmentation outputs that are not only spatially precise but also contextually enriched. By incorporating text-based semantic cues—such as detailed morphological descriptions and explicit boundary characteristics—the method enhances the interpretability of the segmentation results. This dual-modality fusion helps to compensate for the inherent limitations of weak supervision by providing additional context that refines edge delineation and improves overall accuracy across diverse imaging modalities.

\section{Conclusion}
In this paper, we introduced a novel teacher–student framework that leverages two complementary weak supervisory signals—human gaze and vision–language model outputs—to achieve robust medical image segmentation under limited annotation conditions. Our framework first refines sparse yet highly reliable gaze pseudo-masks by fusing them with rich, text-derived morphological and contextual cues in a teacher network, which in turn distills these refined cross-modal features to guide a student network. The student model, trained on more comprehensive but noisier gaze data, is further optimized with confidence-aware consistency loss and a Disagreement-Aware Random Masking strategy that selectively occludes regions with disagreement region, encouraging exploration of less certain areas while reducing reliance on spurious annotations. Extensive experiments on three benchmark datasets (Kvasir-SEG, NCI-ISBI, and ISIC) demonstrate that our approach significantly narrows the performance gap to fully supervised methods while drastically reducing annotation costs. Moreover, the integration of semantic embeddings not only enhances segmentation accuracy but also improves interpretability, which is critical in clinical applications. Despite these strengths, our method has limitations: it relies on the quality and coverage of clinician gaze data and the fidelity of vision–language model outputs—both of which can suffer from noise, domain gaps, or rare pathology appearances—and introduces additional computational complexity in the cross-modal fusion and distillation stages. Clinical deployment also requires further prospective validation and calibration of uncertainty estimates. Future work will extend this framework to additional imaging modalities, other forms of weak supervision (scribbles or automated boxes) and incorporate even richer textual or meta-clinical data, further streamlining annotation processes and bolstering model explainability in real-world medical settings. We also plan to explore active and human-in-the-loop strategies to selectively acquire gaze annotations, methods to better quantify and propagate uncertainty, lightweight fusion architectures to reduce compute overhead, and improved calibration of the language-derived semantic cues.

\section*{Acknowledgment}
The authors gratefully acknowledge Dr. Tao Tan (Faculty of Applied Sciences, Macao Polytechnic University, Macao SAR, China) for thoughtful critiques and helpful discussions that strengthened the presentation and analysis of this paper. This work was supported in part by National Natural Science Foundation of China (62441235) and Beijing Natural Science Foundation (L257005).

\bibliographystyle{IEEEtran}
\bibliography{refs}

\end{document}